\documentclass[conference]{IEEEtran}
\usepackage{blindtext, graphicx}
\usepackage{color}
\usepackage{url}

\usepackage{enumitem}
\usepackage{graphicx}

\begin{document}
%
% paper title
% can use linebreaks \\ within to get better formatting as desired
\title{Social planning for social HRI}
%% CFP https://sites.google.com/site/cogarch4socialhri2016/

% author names and affiliations
% use a multiple column layout for up to three different
% affiliations
\author{\IEEEauthorblockN{Liz Sonenberg\\Tim Miller\\Adrian Pearce}
\IEEEauthorblockA{Department of Computing and\\Information Systems\\
University of Melbourne\\
Australia\\
\emph{Contact: l.sonenberg@unimelb.edu.au}}
\and
\IEEEauthorblockN{Paolo Felli}
\IEEEauthorblockA{University of Nottingham\\UK}
\and
\IEEEauthorblockN{Christian Muise}
\IEEEauthorblockA{MIT, Boston\\USA}
\and
\IEEEauthorblockN{Frank Dignum}
\IEEEauthorblockA{University of Utrecht\\The Netherlands}
}

% conference papers do not typically use \thanks and this command
% is locked out in conference mode. If really needed, such as for
% the acknowledgment of grants, issue a \IEEEoverridecommandlockouts
% after \documentclass

% for over three affiliations, or if they all won't fit within the width
% of the page, use this alternative format:
% 
%\author{\IEEEauthorblockN{Michael Shell\IEEEauthorrefmark{1},
%Homer Simpson\IEEEauthorrefmark{2},
%James Kirk\IEEEauthorrefmark{3}, 
%Montgomery Scott\IEEEauthorrefmark{3} and
%Eldon Tyrell\IEEEauthorrefmark{4}}
%\IEEEauthorblockA{\IEEEauthorrefmark{1}School of Electrical and Computer Engineering\\
%Georgia Institute of Technology,
%Atlanta, Georgia 30332--0250\\ Email: see http://www.michaelshell.org/contact.html}
%\IEEEauthorblockA{\IEEEauthorrefmark{2}Twentieth Century Fox, Springfield, USA\\
%Email: homer@thesimpsons.com}
%\IEEEauthorblockA{\IEEEauthorrefmark{3}Starfleet Academy, San Francisco, California 96678-2391\\
%Telephone: (800) 555--1212, Fax: (888) 555--1212}
%\IEEEauthorblockA{\IEEEauthorrefmark{4}Tyrell Inc., 123 Replicant Street, Los Angeles, California 90210--4321}}

% make the title area
\maketitle

%\begin{abstract}
%something
%\end{abstract}

% IEEEtran.cls defaults to using nonbold math in the Abstract.
% This preserves the distinction between vectors and scalars. However,
% if the journal you are submitting to favors bold math in the abstract,
% then you can use LaTeX's standard command \boldmath at the very start
% of the abstract to achieve this. Many IEEE journals frown on math
% in the abstract anyway.

% Note that keywords are not normally used for peerreview papers.

% % For peer review papers, you can put extra information on the cover
% % page as needed:
% % \ifCLASSOPTIONpeerreview
% % \begin{center} \bfseries EDICS Category: 3-BBND \end{center}
% % \fi
% %
% % For peerreview papers, this IEEEtran command inserts a page break and
% % creates the second title. It will be ignored for other modes.
% \IEEEpeerreviewmaketitle

\section{Research context}

Making a computational agent `social' has implications for how it perceives itself and the environment in which it is situated, including the ability to recognise the behaviours of others at various levels --  simple actions, goals and intentions. Hence  fundamental elements of an architecture for social agents must allow for management of social motivations - i.e. to reach social goals, not only practical goals - and must model and account for actions having both practical and social effects. 
Further it has been argued that to build social agents it is not sufficient to just add a few `social modules' to existing architectures: while multilayer computational cognitive models have been studied for some time, c.f.~\cite{thagard2012}, a new layered deliberation architecture is required that at the higher level(s) naturally accommodates analysis of decision choices that take into account both rich context and future projections of possible consequences, yet does not rely on computational expensive deep reasoning capability~\cite{Dignum2014,Kaminka2013,pacherie2013}. 

In the work reported here, we do not attempt to address the `large' questions associated with the design of a fully integrated computational cognitive architecture; rather we adopt a relatively narrow focus on
exploiting and extending epistemic planning mechanisms to achieve run-time generation of plans in rich multi-actor contexts, i.e. we seek to construct \emph{social plans} in settings where the social context is relevant in the assessment of the beliefs and capabilities of others, and in making appropriate choices of what to do next.

%\section*{Framing remarks}

Our approach has been informed by our experience with the BDI model of agency~\cite{Georgeff1999} and several associated agent architectures - architectures that were introduced to support a balance of deliberative and reactive behaviours, and that in their instantiation are reliant on domain-specific expert knowledge acquisition to provide a \emph{knowledge level} view~\cite{Newell1982}, c.f.~%\cite[p. 5707]{Lemaignan2013}. 
\cite{Lemaignan2013,Norling2008}. 
We are also supporters of the position that logic-based techniques are well suited to represent social reasoning and through which to engineer effective mechanisms, c.f.~\cite{dignum-jasss04,edmonds-jasss04,reich-jasss04}.

Fundamental concepts we build on include: reasoning about the beliefs of others, including their beliefs about others;  establishing common ground;  and the use of stereotypes. So a few words about each.

Exploiting mutual awareness to enable a participant engaged in collaborative activity with others to select an appropriate action typically involves \emph{Theory of Mind} (ToM) reasoning~\cite{Goldman2012,TOM78}, i.e., reasoning about the knowledge, beliefs, perspectives and reasoning of other participants. Agent-based computational models have been used to investigate higher-order ToM in varied scenarios, including alignment with human performance data in some cases~e.g., \cite{Weerd2013,deWeerd2015,Ficici08trails,HiattTraftonHRI15,Malle2015,Thill2015}. 

A specific element of ToM reasoning is \emph{grounding}, or establishing common ground, i.e. an important mechanism by which participants engaged in joint activity coordinate their respective understandings of matters at hand. This construct arises from a model of conversation developed by Herbert Clark~\cite{Wilkes1992} and since studied widely in many fields, including social psychology, e.g.~\cite{Kashima2007}, HCI, e.g.~\cite{Brock1999}, philosophy, e.g.~\cite{Allan2013}. Finding computationally amenable representations and mechanisms that allow agents interacting with humans to keep track of the activity, and their understanding of other participants in the same activity, remains a challenge, c.f.~\cite{pacherie2013,vesper2014support}. Exploring alternative definitions of grounding, allowing for subtle and important variations in the notions of knowledge, belief and acceptance, is one aspect we have investigated~\cite{millercg,Pfau2014s}.

To efficiently take action in settings without the forms of full information needed for ToM reasoning, humans often reason in terms of the (reference) groups to which they and others belong, and the role structures and stereotypical behaviours associated with those reference groups. Steps have been taken towards equipping agents with similar computational capabilities, e.g.,~\cite{felli-ijcai-2015,Pfau2014b,Silverman2012}.

Now to \emph{social planning}. Planning research has for some time yielded highly efficient  mechanisms for plan synthesis  suiting single-agent scenarios.
%, planning for semi-autonomous systems that interact with humans presents significant new challenges.
Input to a planner includes descriptions of the world and effects of available actions, the initial state(s) that the world
might be in before the plan-execution agent performs any actions, and the desired
objectives, such as achieving a goal or performing a specific task.
The output typically consists of either a plan (a sequence of actions
for the agent to perform) or a policy (an action to perform for each state).
However, such descriptions are often insuffient for agents operating in  multi-agent environments. In such environments, a planning agent must consider that other agents have their own actions and mental states, and that these actions and mental states can affect the outcomes and interpretation of its own actions. Thus, such reasoning is inherently a social task.

In environments where an agent needs to plan its
interactions with others, computational complexity increases:
the actions of the other agents can induce a combinatorial
explosion in the number of contingencies to be considered,
making both the search space and the solution size exponentially larger, hence demanding novel methods.
A recent advance is the development of \emph{epistemic planning}~\cite{bolander2011ancl}:
planning according to the knowledge or belief (and iterated knowledge or belief) of other agents, allowing the
specification of a more complex class of planning domains than those mostly concerned with
simple facts about the world.

Building on this work and on recent advances in nondeterministic planning, we have made progress towards the challenge of efficient reasoning both \emph{with} and \emph{about} the incomplete, higher-order, and possibly incorrect beliefs of other individuals as part of the planning process, and how we can plan considering the actions of others.  
Our work involves descriptions and demonstrations-in-use of novel mechanisms for stereotypical and empathetic reasoning, explorations of how this can be used as a theory of mind, and planning while considering the actions of others ~\cite{felli-ijcai-2015,Felli2014ICSR,miller-aaai-16,muise-coin15-teamwork,muise2015aaai,muise-dmap15}.

\section{Challenge Scenarios}

We offer three scenarios that provide challenging settings for social planning.

\section*{Scenario 1}

\noindent This scenario illustrates the need for complex reasoning with others, allowing for possibly limited or faulty perceptions by others of their environment.

\begin{quote}
\em
\small
Consider a self-driving car and a pedestrian each approaching an intersection. A safe plan for each is to wait for the other to go, resulting in a stalemate. With human participants, such encounters are generally resolved with social cues: e.g.\ one signalling to the other using a nod of the head or hand signal. In such cases, cues such as establishing eye contact generate a \emph{common belief} that each party understands who will go first, and each party understands that each understands this, etc.\ 
For a self-driving car to achieve similar interactions with a pedestrian, it will need both sophisticated sensing technology (to accurately recognise the nod or hand-signal) and also rich internal computational mechanisms to  interpret the signal.
However, even physical signals often require social context for their correct interpretation. For example, a young child's inability to correctly assess the belief of others, and therefore, the common belief between themselves and a driver, mean that the driver must consider this when planning its action, and may behave more cautiously.
\end{quote}

% -- topics that are core in multi-agent systems research, but are generally not accompanied by efficiently executable mechanisms. In contrast, exploiting world models to efficiently plan future actions while allowing for non-determinism and uncertainty has been the target of automated planning research, but such approaches do not generally consider the mental states of others. 

\section*{Scenario 2}

\noindent This scenario is inspired
by the Wumpus Hunt and demands 
agents engage in strategic and social reasoning. It has been used to demonstrate the power of theory of mind reasoning~\cite{felli-ijcai-2015,Felli2014ICSR}.

\begin{quote}
\em
\small
The lord of a castle is informed by a peasant that a Wumpus is
dwelling in a dungeon nearby. 
It is known that the Wumpus can be killed by one
hunter alone only if asleep; if awake, two hunters are required.
The lord then tasks the peasant to go to fetch
the White Knight, his loyal champion,  and hunt down the beast
together. The White Knight is known for being irreprehensible,
trustworthy and brave; however, the peasant does not know any knight,
and neither their looks.   
While looking for the White Knight, he runs into the Black
Knight and, believing him the White Knight, tells him about the
quest.
\end{quote}

There is some  additional information that needs to be taken into
account: 
on one hand, the knight knows how a Wumpus can be killed by two hunters,
but he is aware that a simple peasant may get scared by the thought of
confronting an awake Wumpus. 
Also, the peasant can not hunt and is unable to see whether
the Wumpus is awake (he can not approach unnoticed), but the knight can.
Therefore it is not clear to him whether
the peasant can be of any help to the quest.
On the other hand, the knight is aware of the misunderstanding: he knows that the
peasant attributes to him all the good qualities of the
White Knight, so the peasant is confident that the knight won't put
him in danger whenever possible.
%%
%%
%
%The implicit and explicit information of this scenario does not allow a unique understanding of the context, and it is not clear how each agent can use such information. 
%
While on the road, they agree on a protocol: they will enter the dungeon from 
two sides, and the Knight will use a whistle to signal whether the Wumpus is
awake, then they will attack.

\section*{Scenario 3}

\noindent A more difficult challenge problem can be found with the multi-player board game of deception and bluff, \emph{Hattari}~\cite{hattari}. 

\begin{quote}
\em
\small
Hattari involves a  crime scene, three suspects, one victim, and clues. The task is to guess who is the culprit, to accuse him or to deceive the other players! Each player receives a ``suspect profile'' and 5 accusation markers. Three suspect profiles are placed upright in the center of the table, and one profile is placed face down, next to the other three. That is the victim of the crime. The goal is to unmask the the culprit among the three standing suspects. The rules of the game involve selective sharing of information,
but also manipulation of incomplete information among the players, through passing around of pieces as players take turns.
\end{quote}
\begin{center}
\includegraphics[width=0.35\textwidth]{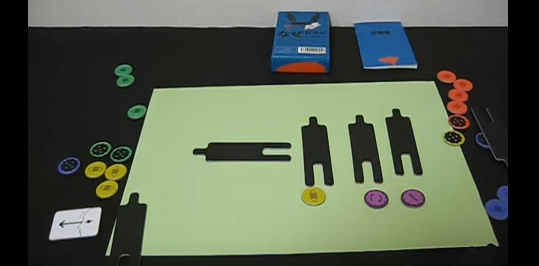}
\end{center}

% https://www.youtube.com/watch?v=CbyMYCiQ79I
% Published on Mar 28, 2013
% This is a little "How to Play" video of the board game Yabu no Naka by Oink Games.
% The latest version of this game is now called Hattari, and there is now a Chinese version of this game also. I haven't seen the English rules for Hattari, so I'm using the original rules I received from Oink Games.
% 0:00 How to Play (basic rules)
% 9:15 How to Play (extra rules)
% 11:13 how to setup for 2/3 players
% 11:50 The box & components

Although we have incorporated some of our research on epistemic planning into a (limited) implementation of  Hattari~\cite{hattari-hac}, creation of an artificial player that could participate meaningfully in a game where humans exploit and interpret body language as they navigate the possibilities of bluffing and deception seems far beyond current technologies.

\section{Workshop discussion questions}
% Contributions are encouraged from all who are interested in participation. A light-touch review process only will be applied to check for relevance, by checking that the questions below are answered in some way - the emphasis of the workshop is on inclusion, discussion and dissemination. Prior to the workshop, the organizers will integrate these into a list of perspectives that will form the basis for the discussions. 
%To facilitate this, we request that all submissions specifically answer the following questions:

%\section*{Responses}

\begin{enumerate}[leftmargin=*]
  \setlength\itemsep{1ex}
\item \emph{Why should you use cognitive architectures - how would they benefit your research as a theoretical framework, a tool and/or a methodology?}

\vspace{1ex}
Our interest is directly in the design of cognitive architectures as the basis for executable strategic collaboration and teamwork involving hybrid human-agent teams. 
%Thus such a cognitive architecture would function as the starting point for laying down the rules that can be used for the reasoning and planning for team work as we are developing \cite{muise-coin15-teamwork}.
\item \emph{Should cognitive architectures for social interaction be inspired and/or limited by models of human cognition?}

\vspace{1ex}
Cognitive architectures should be inspired by models of human cognition. 
Modelling the cognitive architecture after concepts of human cognition seems to allows us to better prepare  agents for human-agent interaction. Further, while explorations with computational models cannot directly shed light on human cognition, c.f.~\cite{Sun2009}, experiments with computational cognitive models can contribute to analyses of potential building blocks for mechanisms involved in coordination in joint action, whether it be in purely human, or human-robot interaction contexts.%, e.g.~\cite{vesper2014support}.
\item \emph{What are the functional requirements for a cognitive architecture to support social interaction?}

\vspace{1ex}
Too many to enumerate here... But, as mentioned above, a cognitive architecture should at least have components modeling the (social) identities, social context and social triggers and effects of actions. In short, representations of the social reality of the partners in the interaction are required.
\item \emph{How the requirements for social interaction would inform your choice of the fundamental computational structures of the architecture (e.g. symbolic, sub-symbolic, hybrid, ...)?}

\vspace{1ex}
Computational structures should be hybrid. For low level interactions and time constrained feedback loops, some very efficient and robust mechanisms are needed that seem best to be represented sub-symbolically. However, for longer term social actions, it is necessary to have symbolic representations in order to deliberate, on the run, about the (social) effects of actions.
\item \emph{What is the primary outstanding challenge in developing and/or applying cognitive architectures to social HRI systems?}

\vspace{1ex}
Outstanding challenges include: identifying and exploiting `sweet spots' in the expressivity-efficiency tradeoff in the engineering of computational artefacts;
finding an effective (domain specific) balance between design-time knowledge engineering and run-time learning; signalling of state (in both directions) between human and artificial participants in joint activity; integration of the diverse perceptual, cognitive and social aspects in a plausibly effective system; establishment of metrics and evaluation methods that allow terms such as ``plausibly effective'' to be precisely defined and formally demonstrated.
\item \emph{Devise a social interaction scenario that current cognitive architectures would likely fail, and why.}

\vspace{1ex}
The beginnings of candidate scenarios are offered above. To provoke failure, what is needed are scenarios exhibiting \emph{social brittleness} - i.e. where the normal course of  interaction fails due to different expectations or assumptions as a result of different social understandings, and a repair has to be found.

\end{enumerate}

\section{Final remarks}

Even though our focus is on cognitive mechanisms as essential components of an integrated cognitive architecture for effective social robots, and we have some exploratory work on human communication patterns~\cite{butchibabu2016}, we recognise there are many topics important in such  architectures that we do not attempt to address -- spatial reasoning~\cite{Warnier2012}, dialogue actions~\cite{Lemaignan2013},  multimodal inputs~\cite{Sridharan2013}, action signalling~\cite{vesper2014support}, the link between perception and action~\cite{Baxter2013,pfau2015iccm}, and
comparisons between logic-based reasoning and other approaches such as game theory~\cite{goodie2012} and probabilistic reasoning~\cite{stuhlmuller2014} ...
to name but a few!!

\section*{Acknowledgements}

Much of the work reported here was carried out while two of the authors (Felli \& Muise) were employed by the University of Melbourne with the financial support of the Australian Research Council Discovery Projects Grant DP130102825 \emph{Foundations of Human-Agent Collaboration: Situation-Relevant Information Sharing}. Additional information about the project can be found at
\url{http://agentlab.cis.unimelb.edu.au/project-hac.html}

% if have a single appendix:
%\appendix[Proof of the Zonklar Equations]
% or
%\appendix  % for no appendix heading
% do not use \section anymore after \appendix, only \section*
% is possibly needed

% use appendices with more than one appendix
% then use \section to start each appendix
% you must declare a \section before using any
% \subsection or using \label (\appendices by itself
% starts a section numbered zero.)
%

% \appendices
% \section{Proof of the First Zonklar Equation}
% \blindtext

% use section* for acknowledgement
% \section*{Acknowledgment}

% The authors would like to thank...

% Can use something like this to put references on a page
% by themselves when using endfloat and the captionsoff option.
% \ifCLASSOPTIONcaptionsoff
%   \newpage
% \fi

% trigger a \newpage just before the given reference
% number - used to balance the columns on the last page
% adjust value as needed - may need to be readjusted if
% the document is modified later
\IEEEtriggeratref{29}
% The "triggered" command can be changed if desired:
%\IEEEtriggercmd{\enlargethispage{-5in}}

% references section

% can use a bibliography generated by BibTeX as a .bbl file
% BibTeX documentation can be easily obtained at:
% http://www.ctan.org/tex-archive/biblio/bibtex/contrib/doc/
% The IEEEtran BibTeX style support page is at:
% http://www.michaelshell.org/tex/ieeetran/bibtex/
\bibliographystyle{IEEEtran}

% for development, use style plain - just easier to work with!!
%%\bibliographystyle{plain}

\bibliography{refs}

\end{document}